\documentclass[sigconf]{acmart}
\renewcommand\footnotetextcopyrightpermission[1]{}
\usepackage{multirow}
\usepackage{amsmath}
\AtBeginDocument{%
  \providecommand\BibTeX{{%
    \normalfont B\kern-0.5em{\scshape i\kern-0.25em b}\kern-0.8em\TeX}}}

\copyrightyear{2019}
\acmYear{2019}
\acmConference[DSHealth 2019]{DSHealth 2019: KDD Workshop on Applied Data Science for Healthcare}{August 05, 2019}{Anchorage, AK}
\acmDOI{00.0000/0000000.0000000}

\acmSubmissionID{123-A56-BU3}

\begin{document}

%

\title[Modeling the Uncertainty with Bayesian Deep Learning]{Modeling the Uncertainty in Electronic Health Records: a Bayesian Deep Learning Approach}

%

\author{Riyi Qiu}
\affiliation{\institution{University of North Carolina at Charlotte}}
\email{rqiu@uncc.edu}

\author{Yugang Jia}
\affiliation{\institution{Boston, United States}}
\email{yugang.jia@gmail.com}

\author{Mirsad Hadzikadic}
\affiliation{\institution{University of North Carolina at Charlotte}}
\email{mirsad@uncc.edu}

\author{Michael Dulin}
\affiliation{\institution{University of North Carolina at Charlotte}}
\email{mdulin3@uncc.edu}

\author{Xi Niu}
\affiliation{\institution{University of North Carolina at Charlotte}}
\email{xniu2@uncc.edu}

\author{Xin Wang}
\affiliation{\institution{Boston, United States}}
\email{wangxin8588@gmail.com}

%
\renewcommand{\shortauthors}{ }

%
\begin{abstract}
Deep learning models have exhibited superior performance in predictive tasks with the explosively increasing Electronic Health Records (EHR). However, due to the lack of transparency, behaviors of deep learning models are difficult to interpret. 
Without trustworthiness, deep learning models will not be able to assist in the real-world decision-making process of healthcare issues. We propose a deep learning model based on Bayesian Neural Networks (BNN) to predict uncertainty induced by data noise. The uncertainty is introduced to provide model predictions with an extra level of confidence. Our experiments verify that instances with high uncertainty are harmful to model performance. Moreover, by investigating the distributions of model prediction and uncertainty, we show that it is possible to identify a group of patients for timely intervention, such that decreasing data noise will benefit more on the prediction accuracy for these patients.  
\end{abstract}

%
%
%
\keywords{Electronic Health Records; Bayesian Neural Network; heteroscedastic uncertainty; data noise; ICU mortality prediction}

%

%
\maketitle

\section{Introduction}

Deep learning has a profound impact on various data-driven applications, such as computer vision, natural language processing, and robotics \cite{collobert2008unified,kendall2017uncertainties,levine2018learning}. The expeditious growth of Electronic Health Records (EHR) is motivating a large number of predictive models to enhance healthcare quality, among which deep learning has achieved significant improvement in performance \cite{cheng2016risk,purushotham2018benchmarking}. EHR is usually sparse and heterogeneous since it contains various longitudinal patient-centered data sources, ranging from diagnosis to healthcare provider information. The feature selection required by conventional machine learning methods can be difficult and necessitate expert knowledge. Deep learning is well known for learning predictive artificial features from raw input, which largely reduces feature engineering efforts \cite{lecun2015deep} and meanwhile distills meaningful information from complicated EHR data.

Despite the promising performance, deep learning has limitations. The black-box structure causes the lack of transparency, hence making it difficult for end users to understand the model's predicting behaviors. To provide reasonable explanations and increase the users' confidence on the results, identifying what the trained model does not know is crucial. 
Uncertainty is a fundamental part of every machine learning phase \cite{ghahramani2015probabilistic,krzywinski2013points}. 
Modeling uncertainty is critical in the cases of ``AI Failure" \cite{synced2018}: the self-driving vehicle can kill pedestrians or the Amazon recruiting tool can be gender or race biased. Similarly, a patient can be falsely recognized as ``low-risk" in the hospital. If high uncertainty was assigned to the wrong predictions, such ``failures" could have been avoided.
Uncertainty can be caused by data noise, such as a blurred pixel or an imputed patient record. 
While making predictions on new data, it is unknown on what characteristics or which part of the data may lead to a better model performance.
Moreover, uncertainty can be introduced from the selection of model structure, such as linear model versus tree-based model. 
Based on different situations, two major types of uncertainty can be concluded \cite{der2009aleatory,gal2016uncertainty}: (1) aleatoric uncertainty, which is caused by the noisy data, such as a wrong observed label or imprecise measurements; (2) the epistemic uncertainty, which includes the uncertainty from both model structure and model parameters. 
Furthermore, aleatoric uncertainty can be divided into homoscedastic uncertainty and heteroscedastic uncertainty. Homoscedastic uncertainty is captured independently of input data, while the heteroscedastic uncertainty is instance-dependent. 
 Bayesian Neural Network (BNN) has been applied to capture the uncertainty in domains such as computer vision and natural language processing \cite{kendall2017uncertainties, gal2016uncertainty, siddhant2018deep}. The key idea of Bayesian approaches is to represent the model weights with some predefined prior distributions and train the model to learn the probability density of the posteriors. Sampling algorithms such as Markov Chain Monte Carlo (MCMC) \cite{neal1993bayesian} or Dropout \cite{gal2016dropout} are then used to draw samples from the distributions for the estimations of model output and uncertainty. Moreover, BNN is usually robust to over-fitting and provides more calibrated models \cite{gal2016theoretically}.

In this paper, we use BNN to calculate the uncertainty, particularly the heteroscedastic aleatoric uncertainty caused by the noisy data from MIMIC-III (Medical Information Mart for Intensive Care) \cite{johnson2016mimic}. Sources of the noise include erroneous entries, sensor degradation, and imputed missing values \cite{purushotham2017benchmark}. Since most of the noise sources are not reflected in the data, we make the data noisier by randomly adding missing values to compare how the uncertainty associates with it. Our contribution is (1) combining the idea of Bayesian learning with Convolutional Neural Network (CNN) models for EHR-related predictive tasks to calculate the heteroscedastic aleatoric uncertainty and to explore the effects of the uncertainty on the deep learning model performance, 
(2) Utilizing the estimated uncertainty to identify a group of patients for whom decreasing data noise will benefit more on the prediction accuracy.

\section{Method}
\subsection{Convolutional Neural Networks for EHR}
Deep Convolutional Neural Networks (CNN) is powerful in image classification \cite{krizhevsky2012imagenet}. It captures local structure of input pixels and learns abstract features. For the prediction tasks on EHR data, the convolution operation can explore local temporal correlations between the patient variables collected at multiple visits and try to capture useful information. Unlike the 2-D convolutions for image processing, 1-D convolutions are often applied over the temporal dimension on EHR data since convolving over the feature dimension is meaningless for clinical use \cite{cheng2016risk}.
\subsection{Bayesian Learning}
To capture heteroscedastic aleatoric uncertainty in a classification model with EHR, we need to estimate the observation noise $\sigma$. In contrast to the homoscedastic uncertainty which assumes constant $\sigma$, the heteroscedastic uncertainty assumes that $\sigma$ is input-dependent \cite{le2005heteroscedastic}. In a normal deep learning model, the network output \textbf{x} is passed into a dense layer with weight \textbf{W} and a Softmax function to predict the probability vector $\hat{p}$: 
\begin{gather*}
     \hat{p} = Softmax(\bf{W}\bf{x}) 
\end{gather*}
Here we add a noise term $k$ to the weight \textbf{W} and place a Gaussian distribution over the $k\textbf{x}$. The standard deviation $\sigma_x$ of the Gaussian distribution is calculated from \textbf{x} by another dense layer:
\begin{gather*}
    \bf{\hat{x}} = (\bf{W}+k)\bf{x} = \bf{W}\bf{x} + k\bf{x},
    \\
    k\bf{x} \sim N(0,\sigma_x^2I)
\end{gather*}
where \textbf{I} stands for an identity matrix. Then, we predict the probability $\hat{p}$ using the ``corrupted" output $\bf{\hat{x}}$ \cite{kendall2017uncertainties}:
\begin{gather*}
    \hat{p} = Softmax(\bf{\hat{x}})   
\end{gather*}
Since there is no analytical solution to integrate out the Gaussian distribution of introduced error \textbf{kx} for a normally used cross entropy loss function for classification tasks in Deep Learning, Monte Carlo (MC) simulation can be used to approximate the objective. The simulation is performed after the calculation of the network output \textbf{x}, so it only increases a fraction of the model computing time. Assume that T times Monte Carlo is simulated, the loss function for this part is:
\begin{gather*}
    \hat{x}_t = \bf{W}\bf{x} + \sigma_x\epsilon_t, \epsilon_t \sim N(0,I)
    \\
    Loss = \log\frac1T\sum_{t=1}^{T}\exp{(\hat{x}_t - \log\sum_{c}\exp{\hat{x}_{t,c}})}
\end{gather*}
where $t$ represents one MC simulation, $c$ is every element in $\hat{x}_t$, and $Loss$ is the Bayesian categorical cross entropy. The Bayesian CNN model will be optimized towards the weighted average of a regularizer on estimated $\sigma_x$ and the categorical cross entropy that are commonly used in normal deep learning classification models. Note that only the classification task is supervised and the aleatoric uncertainty as the variance term $\sigma_x^2$ is learned as we minimize the loss function. 

\section{Experiments}
\subsection{Data Source and Description}
The Medical Information Mart for Intensive Care (MIMIC-III) is a large database consisting of deidentified information related to patients admitted to the intensive care unit (ICU) at the Beth Israel Deaconess Medical Center over 11 years. It contains 38,597 adult patients with 49,785 hospital admissions. We follow the data preprocessing and feature extraction steps described in \cite{harutyunyan2017multitask} to generate the dataset. The task is to predict in-hospital mortality after patient admission to an ICU. The dataset contains a total of 21,139 records, among which 2,797 are positive cases. The observation window is the first 48 hours after admission. Seventeen features are extracted, including heart rate, temperature, weight, pH, Glascow coma scales, and patient monitor records, such as systolic blood pressure and respiratory rate. The categorical features are one-hot encoded and others are normalized. Due to the sparsity of the EHR data, there is a large number of missing values. Each value is imputed and followed by a mask specifying its status (true value or missing value) \cite{harutyunyan2017multitask}. After encoding, normalizing, and imputation, each patient's record is in the shape of 48 hours with 76 generated features. We randomly divide it into training (70\%), validation(15\%), and testing(15\%) sets for the experiments.
\subsection{Implementation}
\subsubsection{Benchmark Model}
We use the 1-D Convolutional Neural Network (CNN) as the benchmark model in the experiments. The network is composed of 5 layers: the input layer, two 1-D convolution layers (each with 50 1x3 convolutions), a layer that deals with over-fitting (dropout, pooling, and normalization), and the output layer that predicts the probability of each class. The optimizer is Adam with the learning rate at 0.001. 
\subsubsection{Bayesian CNN Model}
Our proposed Bayesian CNN model is developed based on the benchmark and uses the same parameter setup. Before the last layer that normal CNN model predicts the probability, the output from previous layers is used for the optimization of the loss function in Section 2. We run 100 times of MC simulations for the approximation of the Gaussian distribution to get smooth and stable results. The weights for the combination of the Bayesian categorical entropy and the normal categorical entropy are 0.2 and 1, respectively. 
\section{Results and Discussion}

\begin{table}
  \caption{Comparison of AUC achieved by Benchmark CNN and Bayesian CNN}
  \label{auccurve1}
  \begin{tabular}{ccl}
    \toprule
    &Benchmark CNN&Bayesian CNN\\
    \midrule
    AUC & $0.8439 \pm 0.0097$ & $0.8507 \pm 0.0116$\\
  \bottomrule
\end{tabular}
\end{table}

\begin{figure}
\captionsetup{width=.49\textwidth}
\centering
\includegraphics[scale=0.55]{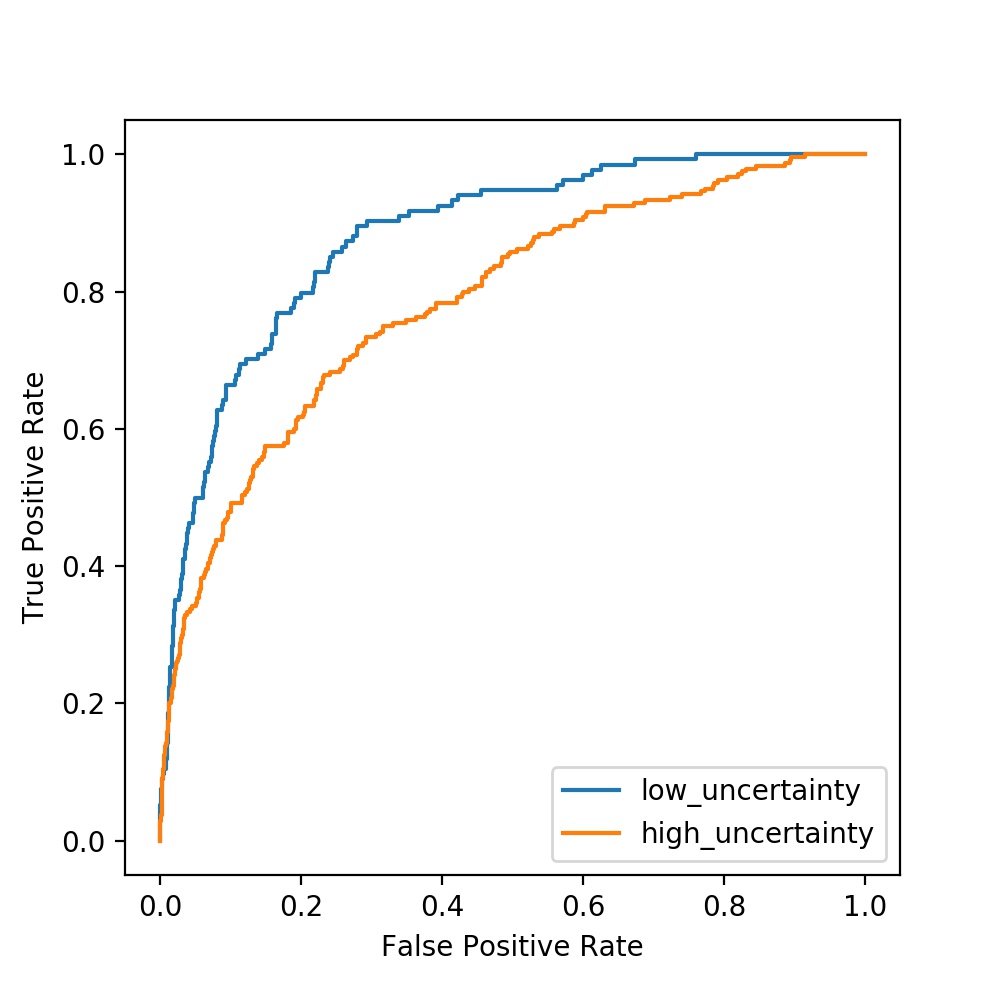}
\caption{The AUC comparison between low uncertainty (0.8593) \& high uncertainty (0.7923) patients.}
\label{auccurve2}
\end{figure}

In this section, we evaluate model performance using the area under the receiver operating characteristic curve (AUC). Table~\ref{auccurve1} exhibits the AUC of Bayesian CNN and the benchmark model, averaged from ten models. The AUCs are close to each other, which demonstrates that predicting uncertainty in the proposed model does not harm its performance. To verify if the uncertainty correlates to the model performance, we also compare the AUC between cases with high uncertainty and cases with low uncertainty. The testing set population is equally divided into two parts at the median of uncertainty distribution. Then, an AUC curve is plotted for each patient group. As shown in Figure~\ref{auccurve2}, cases with low uncertainties tend to outperform the high-uncertainty group. Ten models are randomly trained to verify the observation, and the average AUC increases from $0.7923 \pm 0.0078$ to $0.8593 \pm 0.0170$ when uncertainty moves to the lower part. We compare the low-uncertainty and high-uncertainty patient groups and observe that the average age and percentage of male (or female) are similar, while the high-uncertainty group ($281.2 \pm 17.31$) has more positive cases than low-uncertainty group ($92.8 \pm 17.31$). We believe that it can be explained by the imbalance of the data. Since the data contains a lot more negative cases, the model learns more sufficiently on negative cases than positive cases. Therefore, positive cases tend to present higher uncertainty than negative cases.

\begin{figure}
\captionsetup{width=.49\textwidth}
\centering
\includegraphics[scale=0.55]{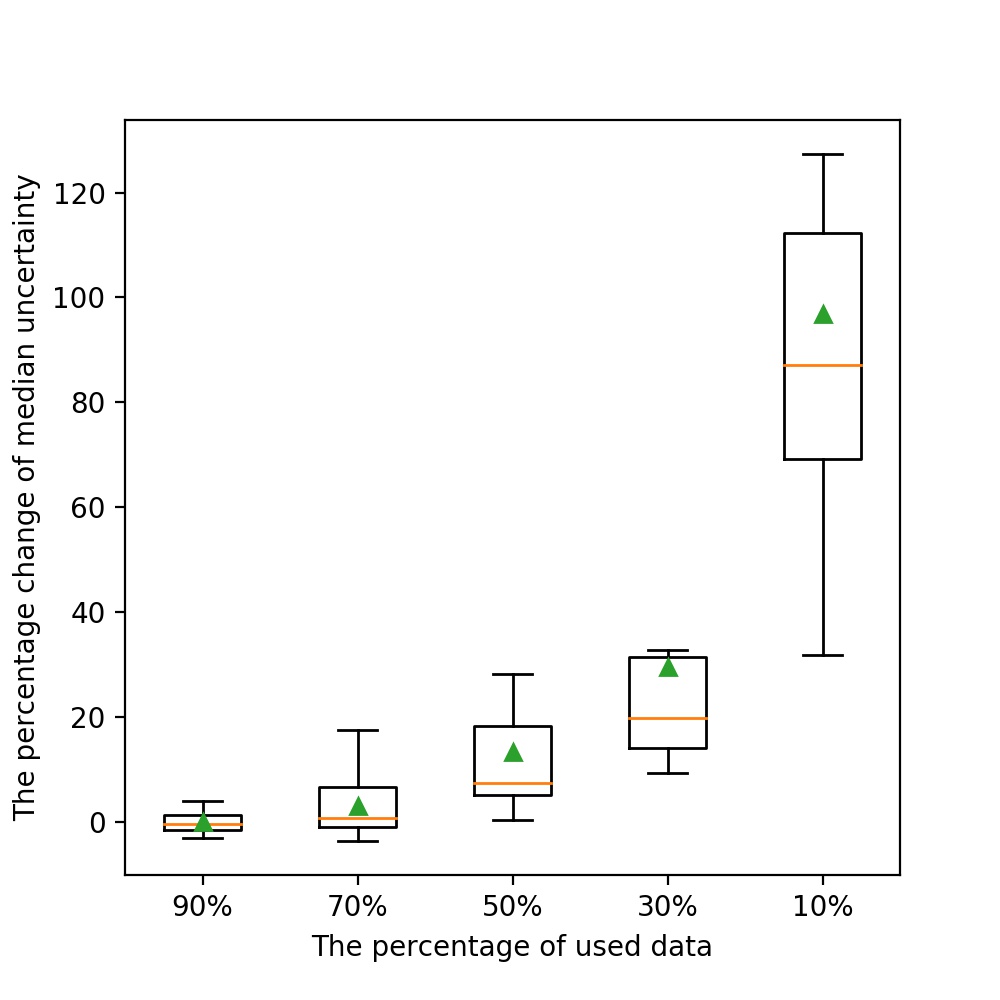}
\caption{The trend of median uncertainty as more data are removed, averaged from 10 models. The orange bar of each plot represents the median; the green triangles represent the means.}
\label{un_change}
\end{figure}

\begin{figure}
\captionsetup{width=.49\textwidth}
\centering
\includegraphics[scale=0.55]{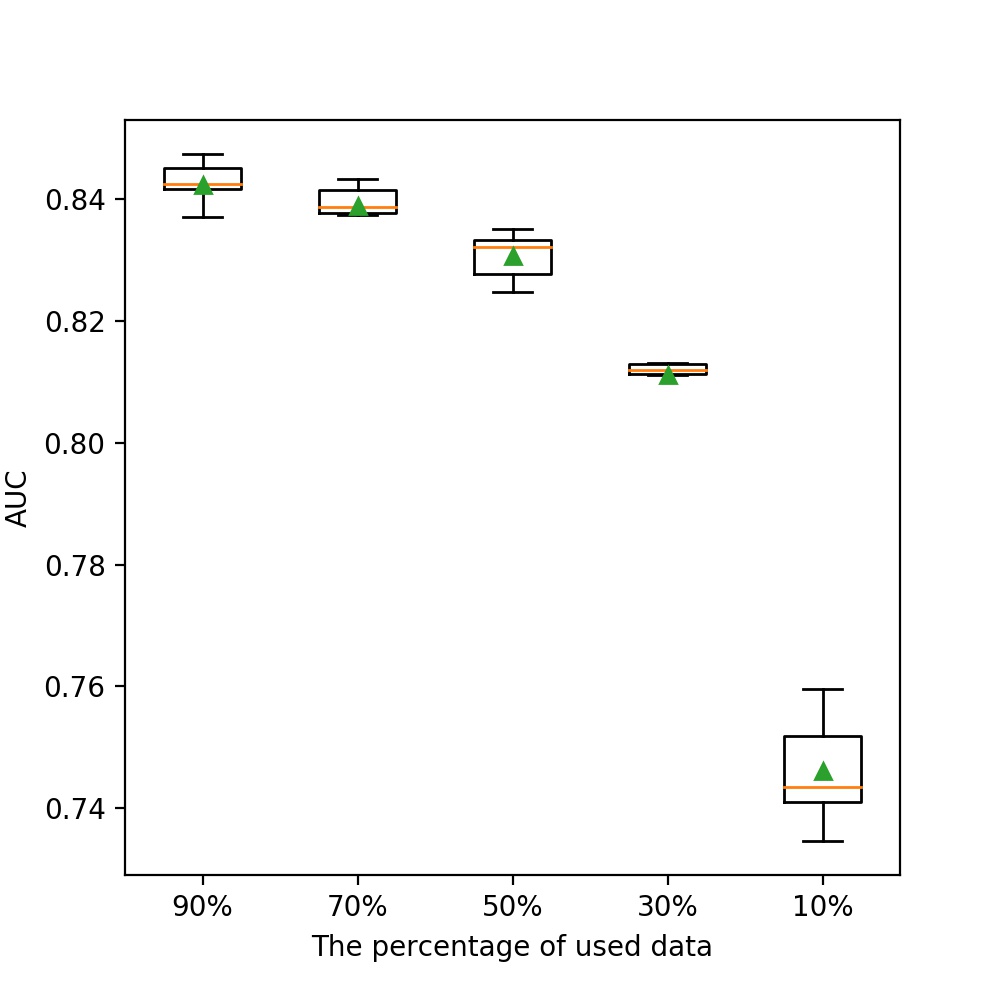}
\caption{The trend of AUC as more data are removed, averaged from 10 models. The orange bar of each plot represents the median; the green triangles represent the means.}
\label{auc_change}
\end{figure}

To verify if data noises that come from missing observations can be captured by the aleatoric uncertainty, we randomly remove part of the raw (before imputation) patient records in the testing dataset. With more missing values imputed, the noise is considered to be higher \cite{cooke2001robust}. Five experiment groups are generated with 90\%, 70\%, 50\%, 30\%, and 10\% of the original data, respectively. Then, the 10 trained models are used to calculate uncertainties and probabilities. As shown in Figure~\ref{un_change}, the box-plot of 10 models indicates that, as more raw data is removed and imputed (higher noise), the median of uncertainty will grow increasingly steeper. Figure~\ref{auc_change} shows the corresponding predicted AUC of each dataset. The trend that AUC decreases is similar to that of increasing uncertainty: with less raw data (higher data noise after the imputation), the model performance tends to drop slowly from 90\% to 30\% but decreases drastically from 30\% to 10\%.

\begin{table}
\caption{\label{QT_50_vs_100_data}The change of half-data-model AUC of each uncertainty\&probability quartile if given the full data.}
\centering
\resizebox{.48\textwidth}{35mm}{
\begin{tabular}{cccc}
\hline
Uncertainty Quartiles & Probability Quartiles & AUC changed(\%)\\
\hline
\multirow{4}{*}{0\% - 25\%} & 0\% - 25\% & $0.0411 \pm 0.1069$\\

& 25\% - 50\% & $0.1004 \pm 0.1658$\\
\
& 50\% - 75\% & $0.1233 \pm 0.1474$\\
\
& 75\% - 100\% & $0.2328 \pm 0.1949$\\
\hline
\multirow{4}{*}{25\% - 50\%} & 0\% - 25\% & $0.1265 \pm 0.1760$\\

& 25\% - 50\% & $0.2451 \pm 0.2064$\\
\
& 50\% - 75\% & $0.3875 \pm 0.2983$\\
\
& 75\% - 100\% & $0.0654 \pm 0.3243$\\
\hline
\multirow{4}{*}{50\% - 75\%} & 0\% - 25\% & $0.1834 \pm 0.2090$\\

& \textbf{25\% - 50\%} & $\bf0.3499 \pm 0.1439$\\
\
& \textbf{50\% - 75\%} & $\bf0.5177 \pm 0.1430$ \\
\
& 75\% - 100\% & $0.1370 \pm 0.2906$\\
\hline
\multirow{4}{*}{75\% - 100\%} & 0\% - 25\% & $0.2651 \pm 0.1422$\\

& \textbf{25\% - 50\%} & $\bf0.4254 \pm 0.2206$\\
\
& \textbf{50\% - 75\%} & $\bf0.4731 \pm 0.1151$\\
\
& 75\% - 100\% & $0.2949 \pm 0.2372$\\
\hline
\end{tabular}}

\end{table}

Furthermore, we are interested in whether the predicted uncertainty can help with the decision making process. Specifically, our hypothesis is that by increasing the data volume (i.e. recording patient features more frequently) for certain patients, is it possible to bring the most improvement in performance to the model? We try to verify this by splitting the testing set into groups and evaluating how they react to the increased data volume. For comparison, the result from the middle bar in Figure~\ref{auc_change} (with 50\% of raw data) is set as the half-data-model AUC. The population of the testing set is equally divided into 16 parts by the distribution of predicted uncertainty and probability (the likelihood of mortality in this work): we first split it into 4 groups by the uncertainty quartile and then divide each quartile into 4 parts by its probability distribution. For each patient group, we predict the probability with full data and use only 50\% data for the prediction of rest patients; with the combined results, we update the AUC and calculate how much it increases from the half-data-model AUC. The calculations are repeated for the 10 trained models, and the results are listed in Table~\ref{QT_50_vs_100_data}. As highlighted in bold, when the uncertainty falls into the third or fourth quartile and the predicted probability falls in the middle (25\% - 75\%), increasing the data volume brings significant improvement to model performance. Therefore, the combination of uncertainty and probability can assist with identifying patients who need more attention and optimizing the distribution of limited medical resources in the ICU.

\section{Conclusions and Future Work}
In this work, we apply the Bayesian CNN model on EHR data to predict ICU mortality and, more importantly, calculate the heteroscedastic aleatoric uncertainty of each prediction. By exploring the relationships between uncertainty, probability, and model performance, we demonstrate that cases with lower uncertainty tend to perform better; we also verify that decreasing the raw data density (as the proxy of increasing data noise) will affect the model performance. Moreover, the results indicate that distributions of uncertainty and probability can potentially help clinicians decide the patients that need to be monitored more frequently. The limitations of this work are: (1) removing part of the raw data does not consider its influence on the future patient monitoring decision and corresponding consequence (death or alive) \cite{gottesman2019guidelines}, and (2) the epistemic uncertainty from model structure and parameters is not considered. Future work should address the confounding factors, explore the epistemic uncertainty, and study how to apply the uncertainty to serve the decision-making process more accurately.
%
\bibliographystyle{ACM-Reference-Format}
\bibliography{acmart}

\end{document}